\documentclass[10pt,twocolumn,letterpaper]{article}

\usepackage{iccv}
\usepackage{times}
\usepackage{epsfig}
\usepackage{graphicx}
\usepackage{amsmath}
\usepackage{amssymb}
\usepackage{booktabs}
\usepackage{comment}

\usepackage[pagebackref=true,breaklinks=true,letterpaper=true,colorlinks,bookmarks=false]{hyperref}

\iccvfinalcopy % *** Uncomment this line for the final submission

\def\iccvPaperID{2} % *** Enter the ICCV Paper ID here

\ificcvfinal\fi

\begin{document}
%% janstuff
\newcommand{\redt}[1]{{\color{red} #1}}
\newcommand{\todo}[1]{[{\bf \color{red} TODO: #1}]}
\newcommand{\jvg}[1]{[{\bf \color{orange} Jan: #1}]}
\newcommand{\slp}[1]{[{\color{magenta} Silvia: #1}]}
\newcommand{\frans}[1]{[{\color{cyan} Frans: #1}]}

\newcommand{\Eqs}[2]{Eqs.~(\ref{eq:#1}-\ref{eq:#2})}
\newcommand{\Eq}[1]{Eq.~(\ref{eq:#1})}
\newcommand{\eq}[1]{\Eq{#1}}
\newcommand{\fig}[1]{Fig.~\ref{fig:#1}}
\newcommand{\tab}[1]{Tab.~\ref{tab:#1}}
\newcommand{\tabs}[2]{Tables~\ref{tab:#1}-\ref{tab:#2}}
\newcommand{\sect}[1]{Section~\ref{sec:#1}}
\newcommand{\chap}[1]{Chapter~\ref{chap:#1}}
\newcommand{\app}[1]{Appendix~\ref{app:#1}}
\newcommand{\alg}[1]{Algorithm~\ref{alg:#1}}
\newcommand{\foot}[1]{$^{\ref{foot:#1}}$}

\def\argmax{\operatornamewithlimits{\rm arg\,max}}

\def\argmin{\operatornamewithlimits{\rm arg\,min}}

\newcommand\eqdef{\mathrel{\overset{\makebox[0pt]{\mbox{\normalfont\tiny\sffamily def}}}{=}}}

\newcommand{\norm}[1]{\left\lVert#1\right\rVert}

% % Choose one of the two follwing:
%\newcommand{\latinphrase}[1]{\textit{#1}}  % always italic
\newcommand{\latinphrase}[1]{\emph{#1}}    % italic in roman text, upshaped in italicized text
%\newcommand{\etal}{\latinphrase{et~al.}\xspace}
%\newcommand{\ie}{\latinphrase{i.e.}\xspace}
%\newcommand{\eg}{\latinphrase{e.g.}\xspace}

%%%%%%%%% TITLE
%\title{Progress Prediction in Surgery Videos}
%\title{How good are video progress prediction methods, really?}
\title{Is there progress in activity progress prediction?}

\author{Frans de Boer$^1$\\
\and
Jan C. van Gemert$^1$\\
\and
Jouke Dijkstra$^2$\\
\and
Silvia L. Pintea$^{1,2}$\\
\and
{\normalsize $^{1}$ Computer Vision Lab, Delft University of Technology}\\
{\normalsize $^{2}$ Division of Image Processing (LKEB), Leiden University Medical Center}\\
}

\maketitle
\ificcvfinal\thispagestyle{empty}\fi

%%%%%%%%% ABSTRACT
\begin{abstract}
Activity progress prediction aims to estimate what percentage of an activity has been completed. 
Currently this is done with machine learning approaches, trained and evaluated on complicated and realistic video datasets. 
The videos in these datasets vary drastically in length and appearance.
And some of the activities have unanticipated developments, making activity progression difficult to estimate. 
In this work, we examine the results obtained by existing progress prediction methods on these datasets.
We find that current progress prediction methods seem not to extract useful visual information for the progress prediction task.
Therefore, these methods fail to exceed simple frame-counting baselines.
We design a precisely controlled dataset for activity progress prediction and on this synthetic dataset we show that the considered methods can make use of the visual information, when this directly relates to the progress prediction.
We conclude that the progress prediction task is ill-posed on the currently used real-world datasets.
Moreover, to fairly measure activity progression we advise to consider a, simple but effective, frame-counting baseline.
\end{abstract}

%%%%%%%%% BODY TEXT
\section{Introduction}
\label{sec:intro} 
Visual activity progress prediction is vital to our day-to-day lives: \eg in cooking, we predict how fast the food is ready; in healthcare, estimating how long a surgery will take allows for better resource allocation and shorter waiting times; and for video-editing knowing where an activity begins and ends helps with automatic cropping of the desired video ranges. 
Here, we define activity progress prediction as the task of predicting the percentage of completion of an activity in a video in an online setting, \ie: without access to the length of the video.   
For our purpose, each video contains a single activity, which covers the complete duration of the video and may consist of multiple phases.
However, we assume there are no phase annotations available, as is generally the case in real-world scenarios. 
The main challenge for progress prediction is extracting meaning from the visual inputs, which, ideally relates to the specific phases of the activity and, thus, enables predicting progress.

To address this challenge, current methods rely on deep networks, such as \textsl{VGG-16} \cite{simonyan2015}, \textsl{ResNet} \cite{he2015}, \textsl{YOLOv2} \cite{redmon2016}, or \textsl{I3D} \cite{carreira2018} to extract visual information. 
Furthermore, to remember information over time, current progress prediction methods \cite{becattini2017,twinanda2019} rely on memory blocks and recurrent connections \cite{hochreiter1997long}.
While these embeddings and recurrent connections are useful for extracting visual information and keeping track of the activity progression over time, they may also overfit to uninformative artifacts. 
Here, we aim to analyze if such undesirable learning strategies are occurring when performing progress prediction.

To this end, we consider the state-of-the-art progress prediction methods \cite{becattini2017,kukleva2019,twinanda2019}, as well as two more simple learning-based methods: a 2$D$-only \textsl{ResNet}, and a \textsl{ResNet} model augmented with recurrent connections. 
We evaluate all these learning methods across three video datasets used for progress prediction: \textsl{UCF101-24} \cite{soomro2012}, \textsl{Breakfast} \cite{kuehne2014, kuehne2016}, and \textsl{Cholec80} \cite{twinanda2016}.
Additionally, we compare the learning-based methods with simple non-learning baseline methods such as simply frame counting.

We evaluate models on various dataset types and regimes. We examine the learning methods when they are presented with the full videos during training. In addition, to avoid overfitting to absolute time\slash frame progression, we also evaluate methods when trained on randomly sampled video segments. 
For randomly sampling video segments, it is not possible to do frame-counting, and only the visual information is available for activity progress prediction.
If the methods should fail to extract useful information from the visual data, they would perform on par with non-learning methods based on frame-counting.
Finally, we design a precisely controlled synthetic progress prediction dataset, \textsl{Progress-bar}, on which the visual information is directly related to the progress.
\begin{figure*}[t!]  
    \centering
    \includegraphics[width=.9\linewidth]{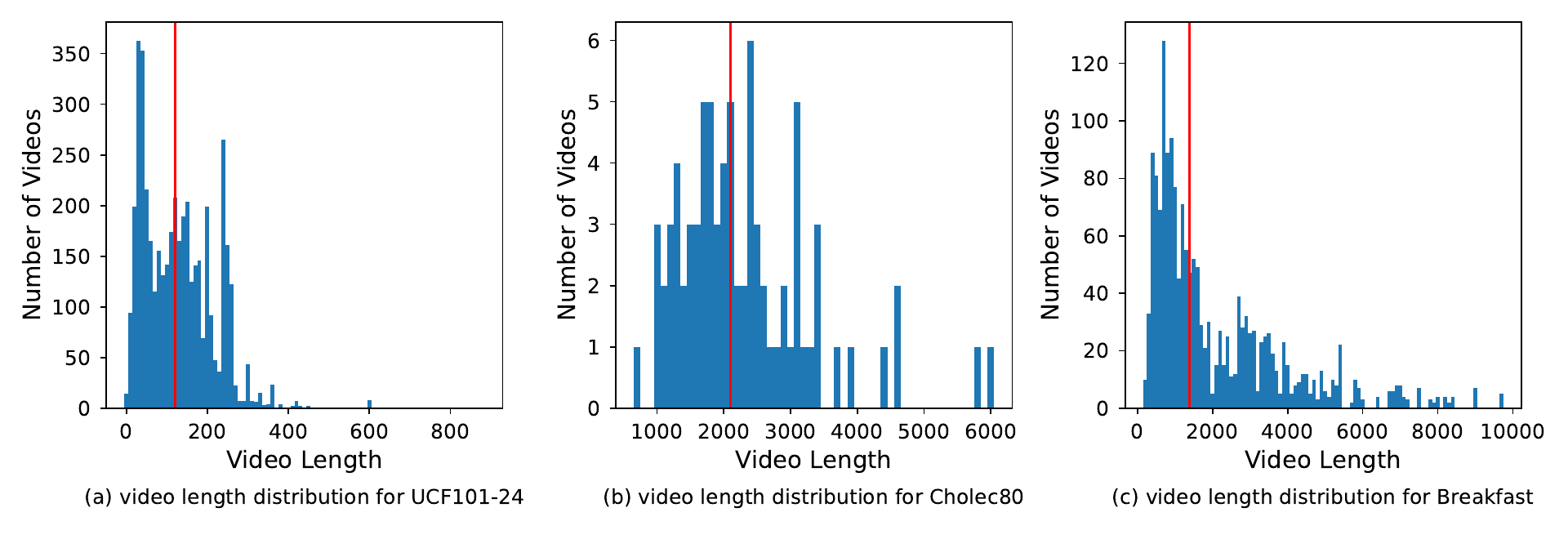}
    \caption{
        Length distributions for \textsl{UCF101-24}, \textsl{Cholec80}, and \textsl{Breakfast}. \textsl{UCF101-24} are grouped into bins of size 10, for \textsl{Cholec80} and \textsl{Breakfast} the bins are of size 100. 
        Most notable is the long-tail distribution of the video lengths in the \textsl{Breakfast} dataset, which makes progress prediction difficult. 
        The vertical red line depicts the mean of each dataset.
   }
    \label{fig:lengths}
\end{figure*}

%-------------------------------------------------------
\newpage
\medskip\noindent\textbf{Difficulties in current progress prediction.}
Progress prediction methods \cite{becattini2017, kukleva2019, twinanda2019} evaluate on complicated and realistic datasets such as \textsl{UCF101-24} \cite{soomro2012}, \textsl{Breakfast} \cite{kuehne2014, kuehne2016}, and \textsl{Cholec80} \cite{twinanda2016}. 
The appearance of the activities in these videos is diverse. 
And the activity length drastically varies between videos in these datasets, as shown \fig{lengths}.
\textsl{UCF101-24} and \textsl{Breakfast} follow a long-tail distribution, with few videos containing long activities.
Moreover, there can be unexpected activity progressions: \eg the pancake gets burned, or there is a surgery lag. 
Also, some of the activities in these datasets do not have a clearly defined endpoint: \eg `skiing', `walking the dog', etc. 
Predicting progress on these activities would be difficult even for a human observer. 
Therefore, we arrive at two main questions we aim to address here: 
(i) \textsl{How well can methods predict activity progression on the current datasets?} and 
(ii) \textsl{Is it at all possible to predict progress from visual data only?}

\section{Related work}
\label{sec:related}

% \cite{becattini2017, hu2019, kukleva2019, han2017, li2017, pucci2023}
% \cite{becattini2017, pucci2023} - closest to there works
% \cite{kukleva2019, li2017} predicts progress, has a downstream task
% \cite{hu2019, han2017} joint prediction task. \cite{han2017} predicts in buckets

\noindent\textbf{Activity progress Prediction.} The task of progress prediction was formally introduced in \cite{becattini2017}. 
Because the progress of an activity is an easy-to-obtain self-supervision signal, it is often used as an auxiliary signal in a multi-task prediction problem, as in \cite{hu2019} to improve the performance of spatio-temporal action localisation networks. 
Progress prediction is also used as a pretext task for phase detection \cite{li2017}, or to create embeddings to perform unsupervised learning of action classes \cite{kukleva2019, vidalmata2020}. 
The progress prediction problem can also be modelled as a classification problem, choosing from $n$ bins each of size $1 / n$ as is done in \cite{han2017}. Based on the literature surveyed, of works done on progress prediction, only \cite{becattini2017, pucci2023} have progress prediction as their primary task. This work is also on the topic of progress prediction, but we do not propose our own progress prediction method. 
Instead, we consider the methods from \cite{becattini2017, kukleva2019} in our analysis and analyze their performance on the currently used datasets.

\smallskip\noindent\textbf{Remaining Duration.} A topic closely related to progress prediction is Remaining Duration (RD) prediction. 
While the goal of progress prediction is to predict the course of the activity as a percentage value in $[0,100\%]$, RD aims at predicting the remaining time $t$ in minutes or seconds. Previous work that researches the RD problem often does this in a surgical setting \cite{aksamentov2017, marafioti2021, twinanda2019, wang2013} and thus refers to it as the Remaining Surgery Duration (RSD) problem. Early methods work by pretraining a \textsl{ResNet-152} model to predict either the surgical phase \cite{aksamentov2017} or the surgery progress \cite{twinanda2019}, and then using the frame embeddings created from the ResNet-152 model in an LSTM block to perform RSD prediction. 
Building on top of this is the observation in \cite{marafioti2021} that predicting extra information such as surgeon skill, may be beneficial to do RSD prediction.
%which shows that by jointly predicting RSD, the surgeon's skill level as either a junior or senior surgeon and the surgical phase the RSD prediction improves. 
Finally, RSD can also be modelled in a way closer to progress. By dividing all RSD values by the highest possible RSD, the RSD can be predicted as a value between $0$ and $1$ \cite{wang2023}. Unlike these methods that model the passage of time as a decreasing remaining duration, we model it as an increasing progress value. 
We use \textsl{RSDNet} \cite{twinanda2019} in our analysis, as it performs both RSD and progress prediction.

\smallskip\noindent\textbf{Phase prediction.} If an action consistently consists of separate sub-tasks or phases of similar duration, then recognizing the current phase gives a good approximation of the progress.  
Previous work jointly performs phase-based progress prediction and surgical gesture recognition \cite{vanamsterdam2020}, jointly predicting the phase and the surgery tools \cite{twinanda2016}, or by using the embeddings in an LSTM to predict the surgical phase online \cite{yengera2018}. 
More recent work applies transformers to perform surgical phase recognition \cite{jamal2023, liu2023lovit}. In this work, we do not consider phase-prediction methods as they are an inaccurate proxy for progress. 
Furthermore, when activities are non-linear, phase prediction is no longer a good indicator of activity progress. 
Knowing which phase is happening may be useful as an extra signal, however we do not consider this, as it requires additional annotations.

\smallskip\noindent\textbf{Activity Completion.} The progress for each frame can be calculated using linear interpolation if the current activity time, $t$, the starting activity time, $t_\text{start}$, and the ending activity time, $t_\text{end}$, are available. Early work on this topic only predicts if an activity has been completed or not using an SVM \cite{heidarivincheh2016}. Follow-up work of Heidarivincheh \etal \cite{heidarivincheh2018} uses a CNN-LSTM architecture to predict the exact frame at which the activity is completed, \ie the activity completion moment. 
The detection of the activity completion moment is done in a supervised setting \cite{heidarivincheh2018}, where the exact frame at which the activity ends is annotated.
Alternatively, activity completion can be done in a weakly supervised setting where the only available annotation is if the activity has been completed or not \cite{heidarivincheh2019}. 
Although related to progress prediction, activity completion only aims at predicting the completion moment. In contrast, we focus on the more fine-grained targets of activity progression at every frame.

\section{Activity progress prediction}
\label{sec:method}
We formulate activity progress prediction as the task of predicting a progress value $p_n^i \in [0, 100]\%$ at frame $i$ in a video indexed by $n$, where
\begin{equation}
  p_n^i = \frac{i}{l_n},
  \label{eq:progress}
\end{equation}
$l_n$ is the total number of frames for video $n$. 
Each video consists of a single activity which starts at frame $1$ and ends at frame $l_n$. 
The activity may consist of multiple phases, but we do not use any phase annotation. 

We predict progress percentages at every frame in the test videos.
During training, the videos can be presented to the methods in two different ways: \textsl{full-videos} and \textsl{video-sequences}. 
We start by using complete videos during training -- \textsl{full-videos}, where each video frame represents a data sample.
Subsequently, we make the problem more realistic by applying two sampling augmentations, as done in \cite{becattini2017}: 
(a) for every video, we sample a segment by randomly selecting a start and end point; 
(b) we randomly subsample every such segment to vary its speed. 
We denote the video sampling strategy implementing both points (a) and (b), as \textsl{video-segments}. 
On \textsl{video-segments} the methods can only rely on the visual information for predicting progress. 

%----------------------------------------------------------------------------------------------------------------
\subsection{Progress prediction methods}
\label{sec:progress_prediction_methods}
We consider 3 progress prediction methods from previous work: \textsl{ProgressNet} \cite{becattini2017}, \textsl{RSDNet} \cite{twinanda2019}, and \textsl{UTE} \cite{kukleva2019}. 
We select these methods as they are the only methods in the surveyed literature that report results on the progress prediction task. 
Furthermore, these methods are the only methods in surveyed literature that do not require additional annotations, such as body joints \cite{pucci2023}.

\smallskip\noindent\textbf{ProgressNet} \cite{becattini2017}: 
A spatio-temporal network which uses a VGG-16 \cite{simonyan2015} backbone to embed video frames and extracts further features using spatial pyramid pooling (SPP) \cite{he2014} and region of interest (ROI) pooling \cite{girshick2015}. 
Additionally, the model uses 2 LSTM layers to incorporate temporal information. 
Becattini \etal also introduce a Boundary Observant (BO) loss. 
This loss enables the network to be more accurate around the areas of phase transitions. 
In our work, we do not use the BO loss because it requires annotating the phase boundaries. 
\textsl{ProgressNet} uses ROI pooling and requires per-frame bounding box annotations. We use the complete frame as the bounding box on datasets where we do not have bounding box annotations.

\smallskip\noindent\textbf{RSDNet} \cite{twinanda2019}: It uses a ResNet-152 \cite{he2015} backbone, followed by an LSTM layer with 512 nodes, and two additional single-node linear layers to jointly predict RSD and video progress. 
The trained ResNet model creates embeddings from all the frames, which are concatenated with the elapsed time in minutes. 
\textsl{RSDNet} jointly trains on RSD and progress prediction but evaluates only on RSD prediction.
Here, we evaluate only the progress prediction head and train with both the RSD and progress loss.

\smallskip\noindent\textbf{UTE} \cite{kukleva2019}: This is a simple 3-layer MLP (Multilayer Perceptron) which takes as input features extracted from RGB video frames such as dense trajectories \cite{wang2013} or I3D network embeddings \cite{carreira2018}. 
Both dense trajectories and I3D embed frames over a sliding window which encodes temporal information into the features. 
This gives the \textsl{UTE} network access to temporal information.
Here, we use 3$D$ convolutional embeddings from the I3D backbone of dimension $1024$ and an embedding window of size $16$ on all datasets. 
We use precisely the same network design as in \cite{kukleva2019}.

%---------------------------------------------------------------------------------------------------------
\subsection{Learning based baselines}
Next to the published methods above, specifically designed for progress prediction, we also consider two more baselines. 
The first is a spatial only \textsl{ResNet-2D} model, and the second is a spatio-temporal \textsl{ResNet-LSTM} model. 
We use \textsl{ResNet-LSTM} as it is a progress-only variation of \textsl{RSDNet}. 
Furthermore, the 2$D$ variant \textsl{ResNet-2D} can give us insights into the spatial-only information contained in the datasets, for progress prediction.
We do not consider other architectures, such as a Video Transformer \cite{arnab2021}, because they do not share the same architecture structure as the progress prediction methods we consider in \sect{progress_prediction_methods}, so they would not display similar behaviors during training.

\smallskip\noindent\textbf{ResNet-2D.} A spatial 2$D$ \textsl{ResNet} \cite{he2015} architecture that can only make use of 2$D$ visual data present in individual video frames, without access to any temporal information. 
The last layer of the \textsl{ResNet} predicts the progress at each frame via a linear layer, followed by a \textsl{sigmoid} activation.

\smallskip\noindent\textbf{ResNet-LSTM.} Additionally, we extend the above \textsl{ResNet-2D} with an LSTM block with 512 nodes, and a final progress-prediction linear layer using a \textsl{sigmoid} activation. 
The LSTM block adds temporal information, which allows us to test the added value of the memory blocks for activity progress prediction. 

%---------------------------------------------------------------------------------------------------------
\subsection{Naive baselines}
Next to the learning-based baselines, we consider a set of naive non-learning baselines. 
These non-learning baselines represent upper-bounds on the errors we expect the learning-based methods to make. 

\smallskip\noindent\textbf{Static-0.5.} This is the most obvious non-learning baseline, which always predicts $50\%$ completion at every frame. This is the best guess without any prior information.

\smallskip\noindent\textbf{Random.} Additionally, we consider a \textsl{random} baseline that predicts a random value in $[0, 100]\%$ at every frame. 
This represents the worst progress prediction a model can make, indicating that it failed to learn anything. 

\smallskip\noindent\textbf{Frame-counting.} Finally, we consider a non-learning baseline which computes training-set statistics.
It is a frame-counting strategy that makes per-frame average progress predictions. 
For frame $i$ in video $n$ this baseline predicts a progress value equal to the average training-progress at frame $i$ of all training videos indexed by $m \in \{1, ..., N_i\}$: 
\begin{equation}
  \hat{p}^i_n = \frac{1}{N_i}\sum_{m=1}^{N_i} p^i_m,
  \label{eq:pf_avg}
\end{equation}
where $N_i$ is the count of all the training videos with a length of at least $i$ frames. 

\section{Empirical analysis}
\label{sec:experiment}
%-----------------------------------------------------------
\subsection{Datasets description}

Each of the considered progress prediction methods evaluates on different datasets: \textsl{RSDNet} on \textsl{Cholec80} \cite{twinanda2016}, \textsl{ProgressNet} on \textsl{UCF101-24} \cite{soomro2012}, and \textsl{UTE} on \textsl{Breakfast} \cite{kuehne2014, kuehne2016}. 
To analyze these methods, we use all 3 datasets for all methods. 

\smallskip\noindent\textbf{\textsl{Cholec80} \cite{twinanda2016}}: Consists of 80 videos of endoscopic cholecystectomy surgery.  
Note that \cite{twinanda2019} uses an extended version of this dataset, \textsl{Cholec120}, containing 40 additional surgery videos. 
However, \textsl{Cholec120} is not publicly available, so we used \textsl{Cholec80} to report our results. 
We randomly create four folds of the data, and follow the same train\slash test dataset split sizes as in \cite{twinanda2019}. 
This dataset has limited visual variability both across training and test splits.
Moreover, the presence of different medical tools in the frames informs of the different surgery phases, which could aid the progress prediction task.

\smallskip\noindent\textbf{\textsl{UCF101-24} \cite{soomro2012}:} Consists of a subset of \textsl{UCF101} containing 24 activities, each provided with a spatio-temporal action tube annotation.\footnote{Following \cite{becattini2017} we use the revised annotations available at \url{https://github.com/gurkirt/corrected-UCF101-Annots}} 
Becattini \etal \cite{becattini2017} split the dataset into 2 categories: \textsl{telic} and \textsl{atelic} activities.
\textsl{Telic} activities are those with a clear endpoint, such as `cliff diving', while \textsl{atelic} activities, such as `walking the dog', do not have a clearly defined endpoint. 
Predicting progress for \textsl{atelic} activities is more difficult than for \textsl{telic} ones.
The original implementation of \textsl{ProgressNet} first trains on \textsl{telic} activities, and then fine-tunes on all activities. 
We did not observe a difference when using this training procedure, and instead train all methods on the full dataset.

\smallskip\noindent\textbf{\textsl{Breakfast} \cite{kuehne2014, kuehne2016}:} Contains 10 cooking activities: \eg `making coffee', `baking pancakes', or `making tea', etc., performed by 52 individuals in 18 different kitchens. 
We use the default splits and train each model across all cooking activities. 
Because the tasks are performed by different individuals in different kitchens, the video appearance varies even within the same task, making this dataset extra challenging for progress prediction.

\textsl{UCF101-24} contains training videos of up to 599 frames, while \textsl{Cholec80} and \textsl{Breakfast} contain videos with thousands of frames.
When training on \textsl{full-videos}, we could not train the \textsl{ProgressNet} model on the original \textsl{Cholec80} and \textsl{Breakfast} datasets, because of the long videos leading to memory problems.
Thus, for the experiments on \textsl{full-videos}, we use a subsampled version of \textsl{Cholec80} from 1 fps to 0.1 fps (the original fps is 25; \cite{twinanda2019} subsamples this down to 1 fps); and we subsample the \textsl{Breakfast} dataset from 15 fps down to 1 fps. 
For our experiments on \textsl{video-segments} we use the original datasets.

\begin{figure*}
\centering
    \begin{tabular}{ccc}
    \multicolumn{3}{c}{\includegraphics[width=.8\linewidth]{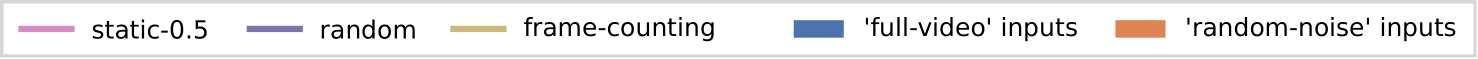}} \\
    \includegraphics[width=0.31\linewidth]{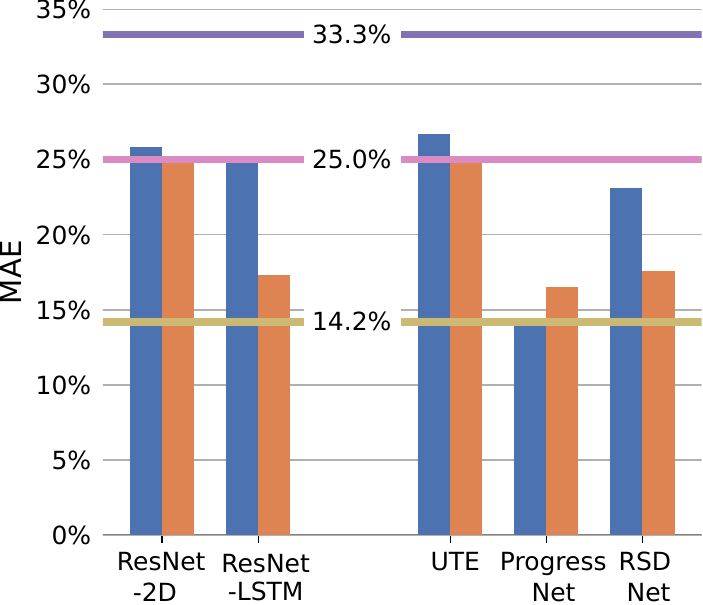} & 
    \includegraphics[width=0.31\linewidth]{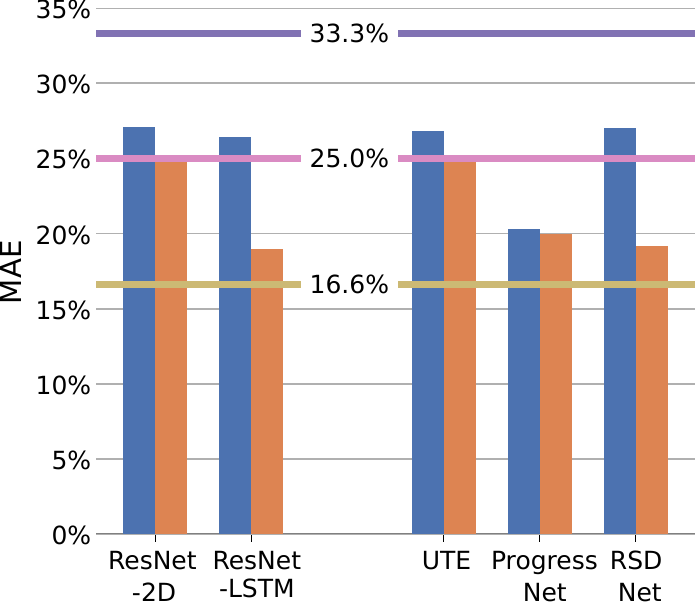} &
    \includegraphics[width=0.31\linewidth]{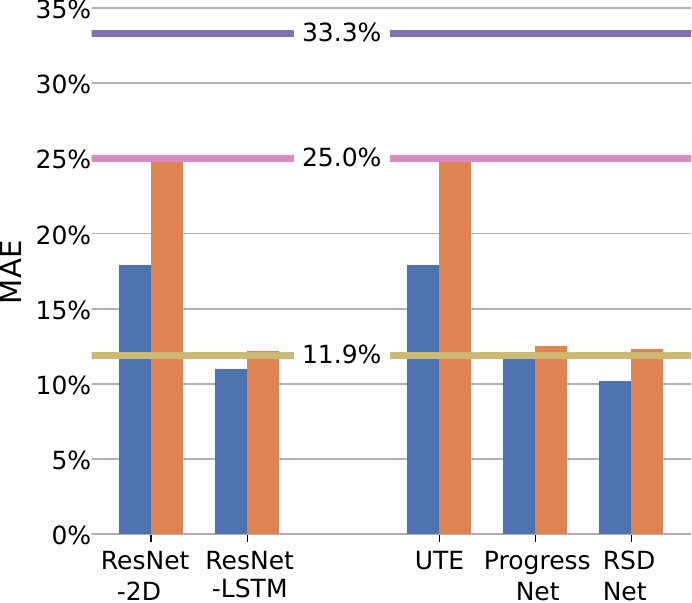} \\
    {\small (a) \textsl{UCF101-24} on \textsl{full-videos}.} &
    {\small (b) \textsl{Breakfast} on \textsl{full-videos}.} &
    {\small (c) \textsl{Cholec80} on \textsl{full-videos}.} \\
    \end{tabular}
   \caption{\textbf{MAE scores on \textsl{full-videos}}. 
       We plot the MAE in percentages for all learning methods when inputting both \textsl{full-video} data and \textsl{random-noise}. 
       (a) MAE for the \textsl{UCF101-24} dataset: For all methods except \textsl{ProgressNet} inputting \textsl{random-noise} performs on par or better than inputting \textsl{full-videos}.
       (b) MAE for the \textsl{Breakfast} dataset: When using \textsl{random-noise} as input to the methods, they perform on par or better than when inputting \textsl{full-videos}, indicating that the methods overfit to training artifacts.
       (c) MAE for the \textsl{Cholec80} dataset: On this dataset, using visual \textsl{full-videos} is better than inputting \textsl{random-noise}, however the \textsl{frame-counting} baseline remains hard to exceed.
   }
\label{fig:full}
\end{figure*}
%-----------------------------------------------------
%-----------------------------------------------------
%-----------------------------------------------------
\subsection{Experimental setup}
For the considered progress prediction methods only the code for \textsl{UTE} is published.\footnote{\url{https://github.com/Annusha/unsup_temp_embed}} 
For the other methods, we follow the papers for implementation details and training procedures. 
We train \textsl{RSDNet} in a 2-step procedure following \cite{twinanda2019}, however for training the LSTM blocks we found that using the Adam optimizer with a learning rate of $10^{-4}$ and no weight decay, for $30$k iterations works best. 
For \textsl{ProgressNet} not all training details are mentioned in the paper, so we use Adam with a learning rate of $10^{-4}$ for $50$k iterations, and we keep the VGG-16 backbone frozen during training. 
For all experiments we report the MAE (Mean Absolute Error) in percentages.
Our code is available online at \href{https://github.com/Frans-db/progress-prediction}{https://github.com/Frans-db/progress-prediction}.

%-----------------------------------------------------
%-----------------------------------------------------
%-----------------------------------------------------
\subsection{\textbf{\emph{(i) How well can methods predict activity progression on the current datasets?}}}
\noindent\textbf{(i.1) Progress predictions on \textsl{full-videos}.}
Here we want to test how well the learning-based models perform when trained on \textsl{full-videos}.
We compare this with using \textsl{random-noise} as input -- we replace each frame with randomly sampled noise. 
Intuitively, learning from \textsl{random-noise} over complete videos will give recurrent models access to frame indices, and this should reach the \textsl{frame-counting} baseline, which computes dataset statistics per frame.
If the models learn to extract useful appearance information, their MAE scores should be considerably higher than when inputting \textsl{random-noise}.
Additionally, we compare the learning-based methods with the naive baselines: \textsl{static-0.5}, \textsl{random}, and \textsl{frame-counting}.

\fig{full}(a) shows that the \textsl{full-video} visual information (blue bars) is less useful than inputting \textsl{random-noise} (orange bars).
When training on the \textsl{full-videos} of \textsl{UCF101-24} both the \textsl{ResNet-2D} and \textsl{UTE} models perform on par with the \textsl{static-0.5} baseline.
This is because these spatial-only networks do not have any way of integrating temporal information and they fail to extract progress information from the visual data alone.
When provided with \textsl{random-noise} as inputs, they always predict $0.5$ and score on par with the \textsl{static-0.5} baseline. 
The results are similar for the recurrent models on \textsl{full-videos}, \textsl{ResNet-LSTM} and \textsl{RSDNet} who are both close to the \textsl{static-0.5} baseline. 
We observe that the recurrent models overfit on the embedded features and fail to generalise. 
When these recurrent networks are provided with \textsl{random-noise} they can only rely on the number of frames seen so far, and thus reach the \textsl{frame-counting} baseline. 
\textsl{ProgressNet} is the only outlier here: when given \textsl{full-videos} it performs better than when given \textsl{random-noise} as input. 
However, \textsl{ProgressNet} still cannot outperform the non-learning \textsl{frame-counting} baseline.

For \textsl{Breakfast} in \fig{full}(b) the results look very similar to those on \textsl{UCF101-24}. 
Both the \textsl{ResNet-2D} and \textsl{UTE} models cannot learn from visual information alone. \textsl{ResNet-LSTM} and \textsl{RSDNet} both perform worse than the \textsl{static-0.5} baseline on \textsl{full-videos}, indicating that they are overfitting on the training data. 
When provided with \textsl{random-noise} as input, they again can only rely on the number of frames seen, and thus approach the \textsl{frame-counting} naive baseline. 

\textsl{Cholec80} in \fig{full}(c) is the only dataset where the spatial-only networks \textsl{ResNet-2D} and \textsl{UTE} perform better than the \textsl{static-0.5} baseline.
Here, we see that using the \textsl{full-videos} (blue bars) is better than inputting \textsl{random-noise} (orange bars).
This hints to the visual information present in this dataset being indicative of the activity progress. 
When inputting \textsl{random-noise} the spatial-only methods again perform on par with the \textsl{static-0.5} baseline, as expected. 
However, this dataset still remains challenging as the methods are not far from the \textsl{frame-counting} baseline.
\textsl{ResNet-LSTM} and \textsl{RSDNet} are the only who perform slightly better than this naive baseline, indicating that they can extract some useful visual information from the video frames. 

\smallskip\noindent\textbf{Observation:} 
\emph{
The current datasets make it difficult for the progress prediction methods to extract useful visual information.
Therefore, the methods overfit to training set artifacts, and are outperformed by simple baselines based on dataset statistics.}

\begin{figure*}
\centering
    \begin{tabular}{ccc}
    \multicolumn{3}{c}{\includegraphics[width=.8\linewidth]{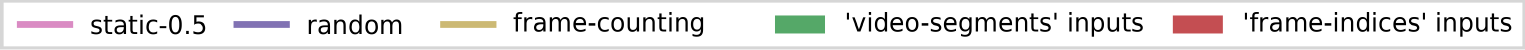}} \\
    \includegraphics[width=0.31\linewidth]{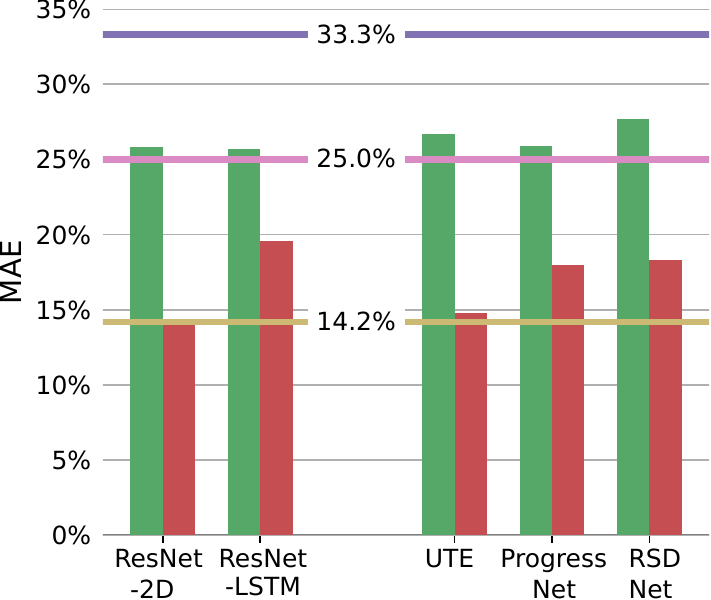} & 
    \includegraphics[width=0.31\linewidth]{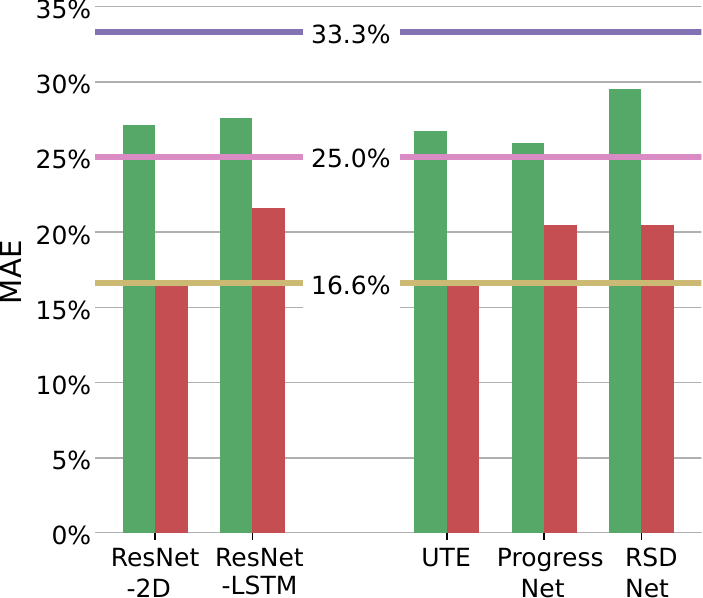} &
    \includegraphics[width=0.31\linewidth]{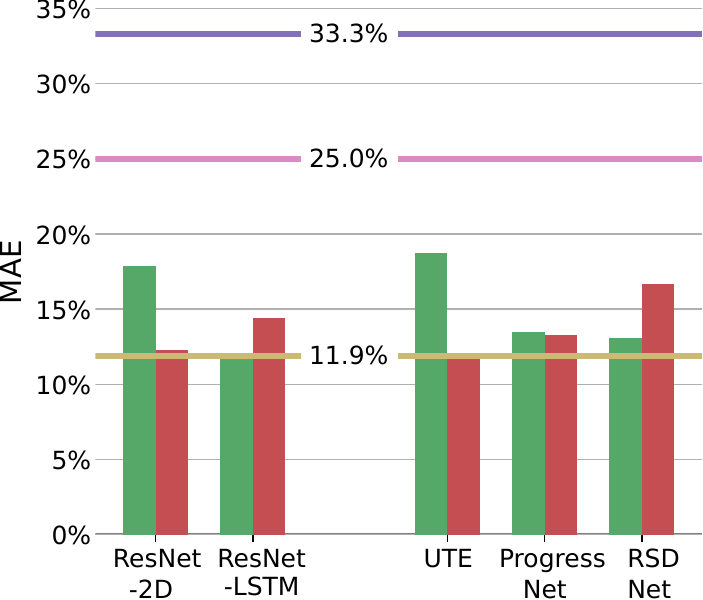} \\
    {\small (a) \textsl{UCF101-24} on \textsl{video-segments}.} &
    {\small (b) \textsl{Breakfast} on \textsl{video-segments}.} &
    {\small (c) \textsl{Cholec80} on \textsl{video-segments}.} \\
    \end{tabular}
   \caption{\textbf{MAE scores on \textsl{video-segments}}. 
       We plot the MAE in percentages for all considered methods when inputting both \textsl{video-segments} and \textsl{frame-indices}. 
       (a) MAE for the \textsl{UCF101-24} dataset: For all methods inputting \textsl{frame-indices} is better than inputting \textsl{video-segments}. 
        \textsl{ResNet-2D} and \textsl{UTE} get the biggest boost in performance because they can learn the one-to-one mapping from index to progress during training. 
       (b) MAE for the \textsl{Breakfast} dataset: Also here \textsl{frame-indices} are more informative than the visual data.
       (c) MAE for the \textsl{Cholec80} dataset: All learning methods perform on par with the \textsl{frame-counting} baseline, except for \textsl{RSDNet} which is slightly worse. 
        This could be due to suboptimal hyperparameter settings. 
   }
\label{fig:seg}
\end{figure*}
%-----------------------------------------------------------------------------------
\medskip\noindent\textbf{(i.2): Progress predictions on \textsl{video-segments}.}
We test the performance of learning methods when trained to predict progress from \textsl{video-segments}.
Using \textsl{video-segments} should encourage the methods to focus more on the visual information and not on the temporal position of the frames, as this is no longer informative. 
We compare inputting \textsl{video-segments} with inputting \textsl{frame-indices} -- the absolute index of each frame, replicated as images.
Intuitively, learning from \textsl{frame-indices} should be on par with the \textsl{frame-counting} baseline, since the only information available is the dataset statistics per frame.
Ideally, we would expect all methods to solve the progress prediction task by relying on visual information, and therefore having lower errors than when inputting \textsl{frame-indices}.
Again, we also compare with the naive baselines: \textsl{static-0.5}, \textsl{random} and \textsl{frame-counting}. 

\fig{seg}(a) shows that the visual information encoded in \textsl{video-segments} (green bars) is less useful than knowing the current frame index (red bars).
When trained on \textsl{video-segments} of \textsl{UCF101-24} all methods perform on par with the \textsl{static-0.5} baseline.
Thus, the models cannot learn to predict progress from the visual video data. 
Interestingly, \textsl{ProgressNet} using \textsl{full-videos} in \fig{full}(a) is better than the \textsl{frame-counting} baseline, however, here it fails to learn when trained on \textsl{video-segments}. 
When provided with \textsl{frame-indices} as input, all methods improve. 
The improvement is most visible for \textsl{ResNet-2D} and \textsl{UTE}, which do not use recurrent blocks. 
This is because the non-recurrent methods can learn the one-to-one mapping from index to progress during training. 

The results on the \textsl{Breakfast} dataset in \fig{seg}(b) are similar to those of \textsl{UCF101-24} in \fig{seg}(a). 
None of the networks can extract useful information from the \textsl{video-segments}.
All methods improve when trained on \textsl{frame-indices}.
The improvement is again more obvious for \textsl{ResNet-2D} and \textsl{UTE}.
Moreover, all results are on par with, or worse than the \textsl{frame-counting} baseline. 

On \textsl{Cholec80} in \fig{seg}(c) all results are close to the \textsl{frame-counting} baseline. 
This is dissimilar to \fig{full}(c) where inputting visual data was better than inputting random noise of the same length as the full video.
Again, \textsl{ResNet-2D} and \textsl{UTE} improve when provided with \textsl{frame-indices} as input. 
For \textsl{ResNet-LSTM} and \textsl{ProgressNet} and \textsl{RSDNet} the performance is on par with the \textsl{frame-counting} baseline when trained on \textsl{video-sequences} indicating that these methods overfit to the training data.
When trained on \textsl{frame-indices} most methods approach the \textsl{frame-counting} baseline, as this is the information encoded in the frame indices across the full training set.
\textsl{RSDNet} performs worst when given \textsl{frame-indices} as inputs;  we hypothesise that this is due suboptimal hyperparameter settings. 

\smallskip\noindent\textbf{Observation:} 
\emph{
When restricting the models to rely only on visual information, the models are outperformed by simply considering the current frame index, and performing dataset statistics.
This is due to the current progress prediction datasets not containing sufficient visual information to guide progress prediction.}

%-------------------------------------------------------------------------------
\subsection{\emph{\textbf{(ii) Is it at all possible to predict activity progress from visual data only?}}}
We observe that current progress prediction datasets are not well-suited for the task, as the learning models fail to extract useful information from visual data alone. 
To test that the problem is indeed with the datasets used for evaluation and not the learning models, we test here if progress prediction is possible from visual data alone.
For this we aim to construct a synthetic dataset in such a way that the learning-based methods perform optimal using visual information, and outperform all the naive baselines.

\begin{figure}[t]
\begin{center}
   \includegraphics[width=1.0\linewidth]{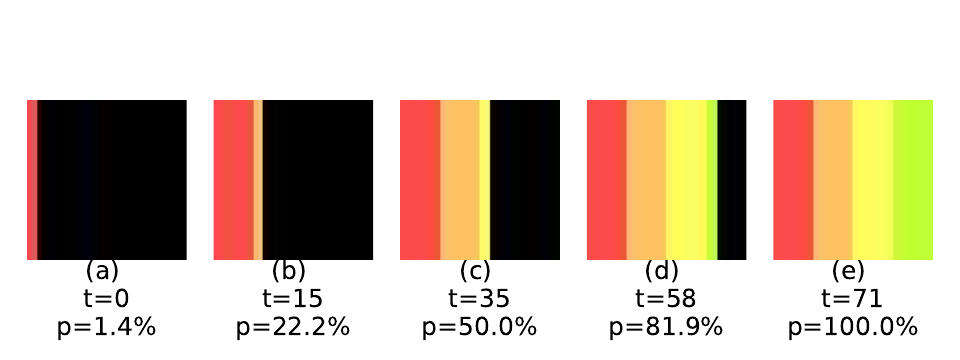}
\end{center}
   \caption{Visualisation of a progress bar from our synthetic \textsl{Progress-bar} dataset at timesteps $t{=}0$, $t{=}15$, $t{=}35$, $t{=}58$, and $t{=}71$. 
   Each coloured section indicates visually a 25\% section, but due to variance in the speed, the actual video progress may differ at these points.
   }
\label{fig:progressbars}
\end{figure}

Our synthetic \textsl{Progress-bar} dataset, shown in \fig{progressbars}, contains a progress bar (similar, for example, to a download bar) that slowly fills up from left to right. 
Each frame has a resolution of $32{\times}32$px. 
We generate $900$ videos for the training split, and $100$ for the test split.
Each bar has its own rate of progression, but there is a variance per notch causing some uncertainty. 
Therefore, even on this simple dataset it is impossible to make no errors.
This is why in the first image the progression appears to be slightly beyond 25\%, but because the video may slow down after this section it is actually at 22.2\%. 
Due to the large variance in video length, ranging from $19$ to $145$ frames, the \textsl{frame-counting} baseline, and thus frame-counting strategies, will give worse results than relying on visual information.
Also, because of the different progress rates per video, the learning methods cannot just rely on visual information alone, but also have to use temporal information to perform well on this progress prediction task. 
Due to the reduced frame resolution and data complexity of our synthetic dataset, we scale down the \textsl{ResNet} backbone, for these experiments. 
Specifically, to avoid overfitting, we use \textsl{ResNet-18} as a backbone for \textsl{ResNet-2D}, \textsl{ResNet-LSTM}, and \textsl{RSDNet}. 
\textsl{ProgressNet} and \textsl{UTE} remain unchanged.

\begin{figure}
\begin{center}
   \includegraphics[width=1.0\linewidth]{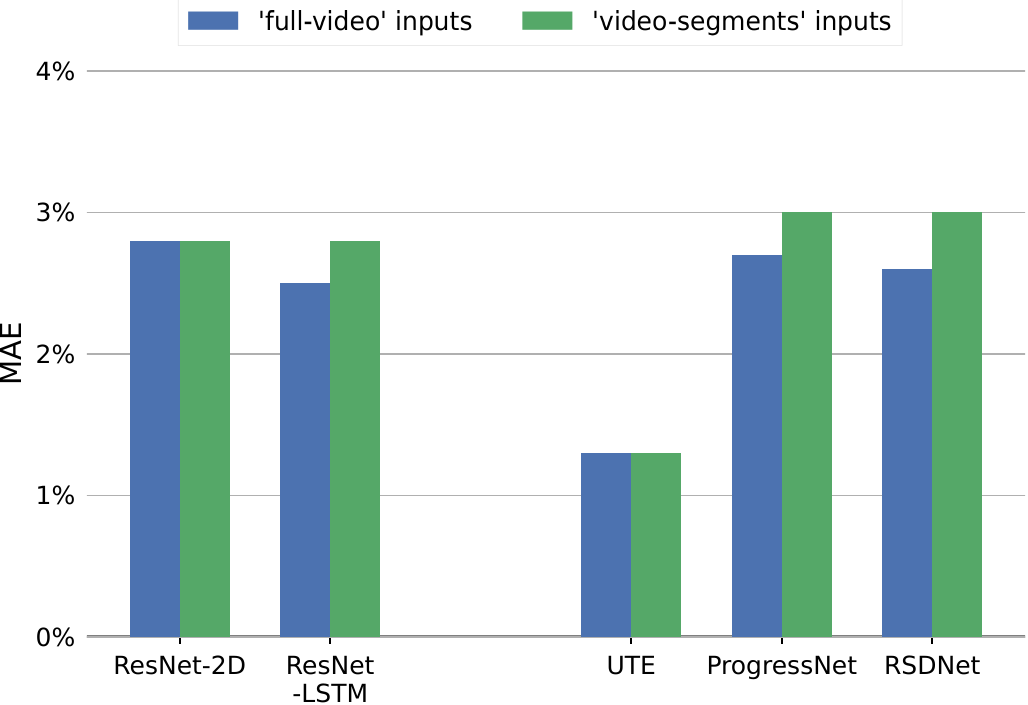}
\end{center}
   \caption{\textbf{MAE scores on our synthetic \textsl{Progress-bar} dataset, when training on \textsl{full-videos} and \textsl{video-segments}}. 
   The \textsl{frame-counting} baseline has an MAE of $12.9\%$, while the \textsl{static} baseline is at $25\%$ and the \textsl{random} baseline at $33.3\%$. 
   We see that all methods outperform the \textsl{frame-counting} baseline. 
   \textsl{UTE} obtains the best result due to its $15$-frame temporal window, which allows it to see $7$ frames into the future. 
   We conclude that the progress prediction methods are able to learn progress from visual information, if it is clearly present in the videos.}
\label{fig:result_bars}
\end{figure}

\fig{result_bars} shows the results of all the learning-based methods when predicting progress from both \textsl{full-videos} and \textsl{video-segments}. 
For this dataset the \textsl{frame-counting} baseline has an MAE of $12.9$\%, which is outperformed by all learning-based methods. 
\textsl{UTE} performs the best out of all the networks, even though it does not have memory. 
This is because \textsl{UTE} relies on 3D convolutional embeddings over a temporal-window of size $16$ frames. 
This temporal-window gives the method information about $7$ future frames, which is sufficient on this simple dataset. 
For the LSTM-based methods inputting \textsl{full-videos} still performs slightly better than inputting \textsl{video-segments}. 
At a closer look, this is because the \textsl{video-segments} sampling method has a bias towards frames in the middle of a video. 
The earlier frames are less likely to get sampled, thus the progress prediction methods will have a higher error there.

\smallskip\noindent\textbf{Observation:} 
\emph{It is feasible for the progress prediction methods to make effective use of the visual data present in the videos and outperform the \textsl{frame-counting} baseline, when the visual data is a clear indicator for the video progression.
}

\section{Discussion and limitations of our analysis}

\begin{figure*}
\begin{center}
    \begin{tabular}{cc}
    \includegraphics[width=.45\linewidth]{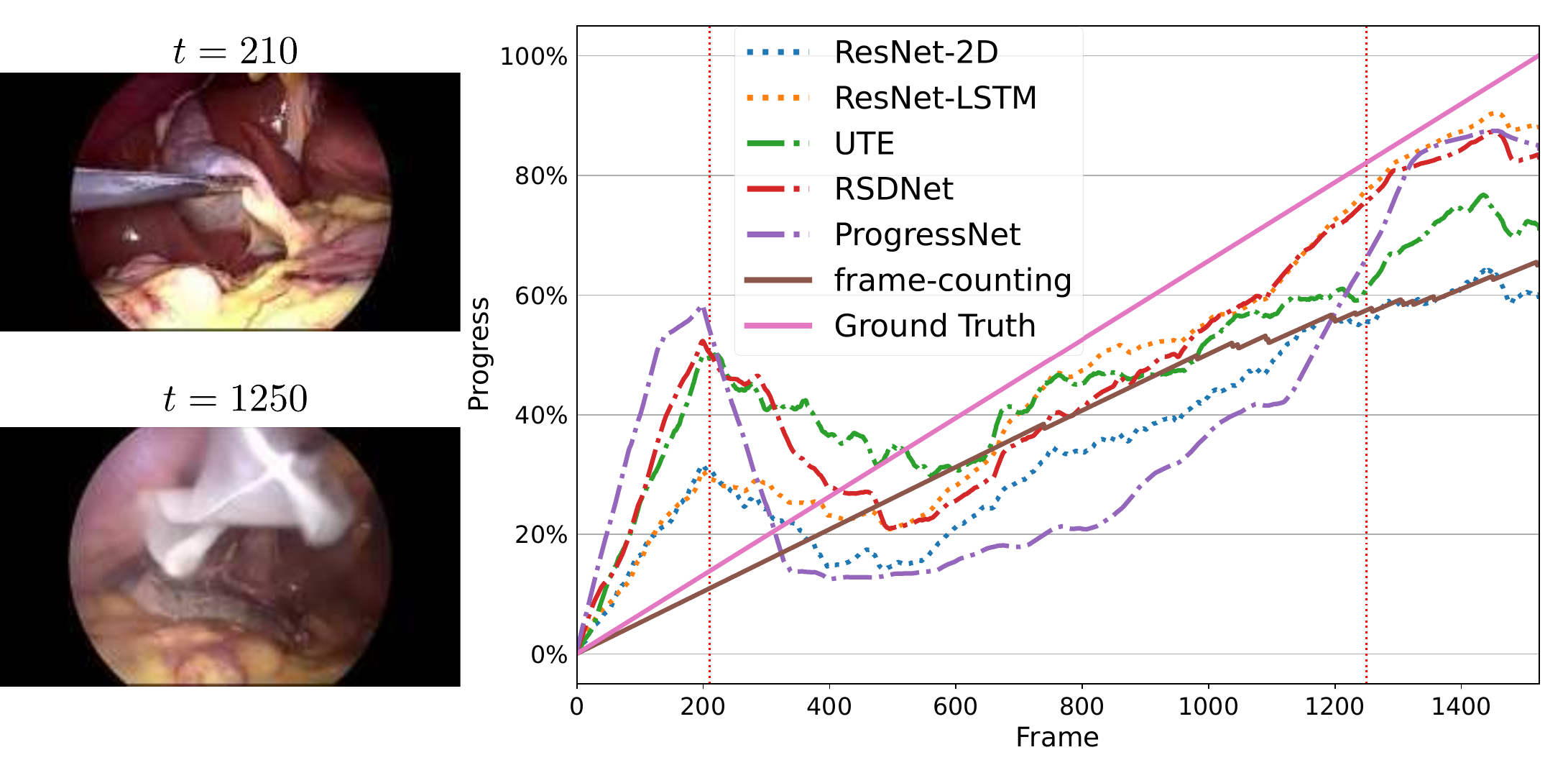}
    & \includegraphics[width=.45\linewidth]{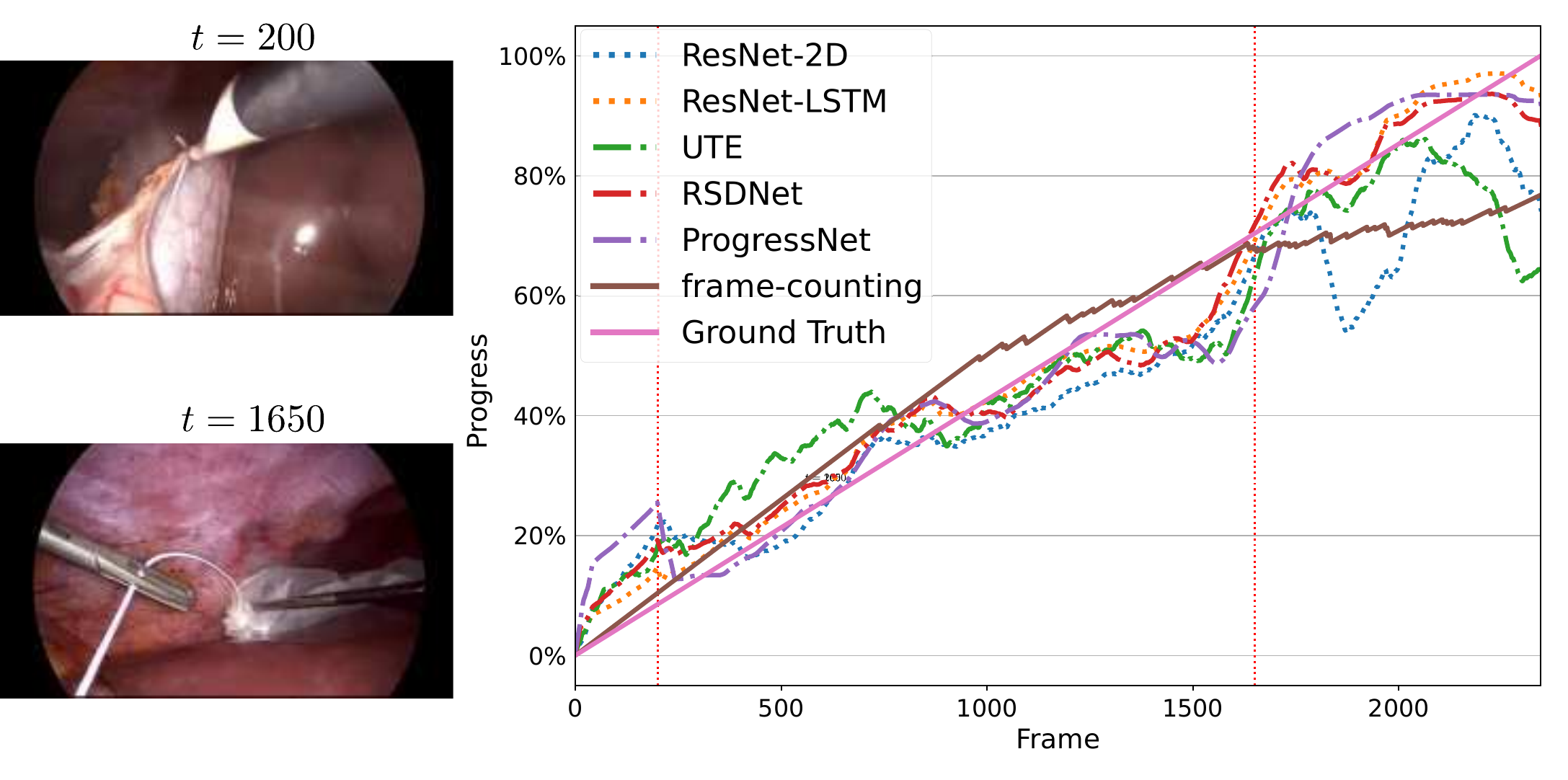} \\
    {\small (a) Video-04 of \textsl{Cholec80}} & 
    {\small (b) Video-05 of \textsl{Cholec80}} \\
    \end{tabular}
\end{center}
   \caption{Activity progress prediction examples of \textsl{Cholec80}.
   (a) Video-04 at timestamps $t{=}210$ and $t{=}1250$. 
   (b) Video-05 at timestamps $t{=}200$ and $t{=}1650$. 
    At $t{=}210$ and $t{=}200$ the methods recognize the medical tool, and correct their progress downwards to signal the start of the medical procedure.
    At $t{=}1250$ and $t{=}1650$ the methods recognize the collection bag and correct their progress to signal the end of the procedure.}
\label{fig:cholec_videos}
\end{figure*}

\begin{figure}
\begin{center}
    \includegraphics[width=1.0\linewidth]{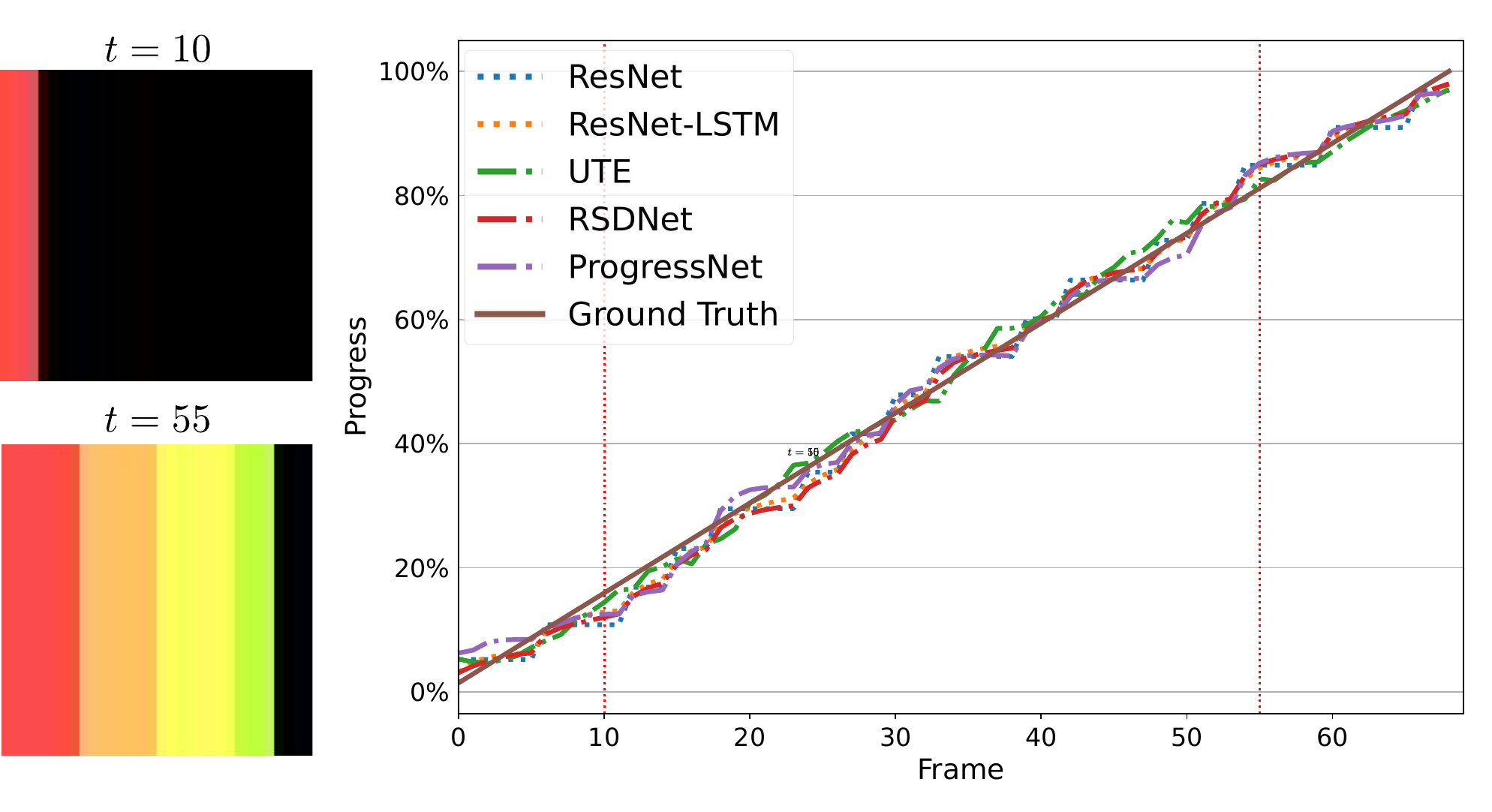} 
\end{center}
   \caption{Progress prediction example on Video-00015 of our synthetic \textsl{Progress-bar} dataset at timestamps $t{=}10$ and $t{=}55$. 
   The learning methods can almost perfectly follow the ground truth.}
\label{fig:bars_00015}
\end{figure}

\noindent\textbf{Discussion.} 
This paper empirically shows that the current progress prediction datasets do now allow for learning useful visual information, and methods are outperformed by naive baselines relying on dataset statistics.
We observe that the progress prediction models perform well on the training data, yet cannot generalize well to the unseen test data.
As future research, it would be interesting to pinpoint the photometric artifacts that the models overfit to.
However, we also saw that some useful visual information was learned on the \textsl{Cholec80}.
This may be due to the presence of clear visual phase delineators. 
\fig{cholec_videos}(a) and \fig{cholec_videos}(b) show examples of predictions on the \textsl{Cholec80} dataset. 
The first frames highlighted in these videos ($t{=}210$ and $t{=}200$) are the moment when the first medical tool is present in the video, and the progress prediction methods adjust their predictions to this new visual information. 
Similarly, the second highlighted timesteps ($t{=}1250$ and $t{=}1650$) represent the moment the collection bag is present which signals the end of the procedure. 
On our synthetic \textsl{Progress-bar} dataset in \fig{bars_00015} we also show the predictions and highlight two moments in the videos. 
Here, the networks almost perfectly follow the ground truth progression. 
These results illustrate that for progress prediction is essential to have clearly recognizable visual transition points, that consistently correspond to a certain progress prediction percentage. 
This is related to the idea of Becattini \etal \cite{becattini2017} who use phase annotations to increase the loss around the phase boundaries.

\smallskip\noindent\textbf{Limitations.} 
The first limitation of our research is that we could only find 3 progress prediction methods to analyze, on 3 datasets. 
Additionally, we do not consider here other video-architectures such as a Video Transformer \cite{arnab2021}, as these are not directly related to the progress prediction methods we analyze.
However, we do consider 2D (ResNet) and 3D (I3D) convolutional embeddings, as well as recurrent networks (with LSTM blocks). 
Thirdly, we were unable to match the results of \textsl{ProgressNet} exactly as reported in \cite{becattini2017}: 
when trained on \textsl{video-segments}, the authors report an MSE of $0.052$ (MAE of approximately $22.8$\%), while we obtain an MAE of $25.9$\%.
Nonetheless, the \textsl{frame-counting} outperforms the result reported in \cite{becattini2017}, which still validates our conclusions.
Finally, we observed that on both \textsl{UCF101-24} and \textsl{Breakfast} the methods have a tendency to overfit. 
Maybe better strategies to overcome this overfitting phenomenon could improve the results.

\section{Conclusion}
\label{sec:conclusion}
In this paper, we investigate the behaviour of current progress prediction methods on the currently used benchmark datasets. 
We show that on the currently used datasets, the progress prediction methods can fail to extract useful information from visual data, and are exceeded by simple non-learning baselines based on frame counting. 
Additionally, we evaluate all the methods on a synthetic dataset we specifically design for the progress prediction task. 
On our synthetic dataset the results show that all the methods can make use of the visual information and outperform the naive, non-learning baselines. 
We conclude that in its current form the task of progress prediction is ill-posed, as the currently used datasets for progress prediction are not suitable for this task.

\small
\smallskip\noindent\textbf{Acknowledgements.} 
This work was done with the support of the Eureka cluster Program, IWISH project, grant number AI2021-066.
Jan van Gemert is financed by the Dutch Research Council (NWO) (project VI.Vidi.192.100).

{\small
\bibliographystyle{ieee_fullname}
\bibliography{egbib}

\begin{thebibliography}{10}\itemsep=-1pt

\bibitem{aksamentov2017}
Ivan Aksamentov, Andru~Putra Twinanda, Didier Mutter, Jacques Marescaux, and
  Nicolas Padoy.
\newblock Deep neural networks predict remaining surgery duration from
  cholecystectomy videos.
\newblock In {\em International Conference on Medical Image Computing and
  Computer-Assisted Intervention}, 2017.

\bibitem{arnab2021}
Anurag Arnab, Mostafa Dehghani, Georg Heigold, Chen Sun, Mario Lučić, and
  Cordelia Schmid.
\newblock Vivit: A video vision transformer, 2021.

\bibitem{becattini2017}
Federico Becattini, Tiberio Uricchio, Lorenzo Seidenari, Lamberto Ballan, and
  Alberto Del~Bimbo.
\newblock Am i done? predicting action progress in videos, 2017.

\bibitem{carreira2018}
Joao Carreira and Andrew Zisserman.
\newblock Quo vadis, action recognition? a new model and the kinetics dataset,
  2018.

\bibitem{heidarivincheh2016}
Majid~Mirmehdi Farnoosh~Heidarivincheh and Dima Damen.
\newblock Beyond action recognition: Action completion in rgb-d data.
\newblock In Edwin R.~Hancock Richard C.~Wilson and William A.~P. Smith,
  editors, {\em Proceedings of the British Machine Vision Conference (BMVC)},
  pages 142.1--142.11. BMVA Press, September 2016.

\bibitem{girshick2015}
Ross Girshick.
\newblock Fast r-cnn, 2015.

\bibitem{han2017}
Tengda Han, Jue Wang, Anoop Cherian, and Stephen Gould.
\newblock Human action forecasting by learning task grammars, 2017.

\bibitem{he2014}
Kaiming He, Xiangyu Zhang, Shaoqing Ren, and Jian Sun.
\newblock Spatial pyramid pooling in deep convolutional networks for visual
  recognition.
\newblock In {\em Computer Vision {\textendash} {ECCV} 2014}, pages 346--361.
  Springer International Publishing, 2014.

\bibitem{he2015}
Kaiming He, Xiangyu Zhang, Shaoqing Ren, and Jian Sun.
\newblock Deep residual learning for image recognition, 2015.

\bibitem{heidarivincheh2018}
Farnoosh Heidarivincheh, Majid Mirmehdi, and Dima Damen.
\newblock Action completion: A temporal model for moment detection, 2018.

\bibitem{heidarivincheh2019}
Farnoosh Heidarivincheh, Majid Mirmehdi, and Dima Damen.
\newblock Weakly-supervised completion moment detection using temporal
  attention.
\newblock In {\em 2019 IEEE/CVF International Conference on Computer Vision
  Workshop (ICCVW)}, pages 1188--1196, 2019.

\bibitem{hochreiter1997long}
Sepp Hochreiter and J{\"u}rgen Schmidhuber.
\newblock Long short-term memory.
\newblock {\em Neural computation}, 9(8):1735--1780, 1997.

\bibitem{hu2019}
Bo Hu, Jianfei Cai, Tat-Jen Cham, and Junsong Yuan.
\newblock Progress regression rnn for online spatial-temporal action
  localization in unconstrained videos, 2019.

\bibitem{jamal2023}
Muhammad~Abdullah Jamal and Omid Mohareri.
\newblock Surgmae: Masked autoencoders for long surgical video analysis, 2023.

\bibitem{kuehne2014}
Hilde Kuehne, A.~B. Arslan, and T. Serre.
\newblock The language of actions: Recovering the syntax and semantics of
  goal-directed human activities.
\newblock In {\em Proceedings of Computer Vision and Pattern Recognition
  Conference (CVPR)}, 2014.

\bibitem{kuehne2016}
Hilde Kuehne, Juergen Gall, and Thomas Serre.
\newblock An end-to-end generative framework for video segmentation and
  recognition.
\newblock In {\em Proc. IEEE Winter Applications of Computer Vision Conference
  (WACV 16)}, Lake Placid, Mar 2016.

\bibitem{kukleva2019}
Anna Kukleva, Hilde Kuehne, Fadime Sener, and Juergen Gall.
\newblock Unsupervised learning of action classes with continuous temporal
  embedding, 2019.

\bibitem{li2017}
Xinyu Li, Yanyi Zhang, Jianyu Zhang, Yueyang Chen, Shuhong Chen, Yue Gu,
  Moliang Zhou, Richard~A. Farneth, Ivan Marsic, and Randall~S. Burd.
\newblock Progress estimation and phase detection for sequential processes,
  2017.

\bibitem{liu2023lovit}
Yang Liu, Maxence Boels, Luis~C. Garcia-Peraza-Herrera, Tom Vercauteren, Prokar
  Dasgupta, Alejandro Granados, and Sebastien Ourselin.
\newblock Lovit: Long video transformer for surgical phase recognition, 2023.

\bibitem{marafioti2021}
Andrés Marafioti, Michel Hayoz, Mathias Gallardo, Pablo~Márquez Neila,
  Sebastian Wolf, Martin Zinkernagel, and Raphael Sznitman.
\newblock Catanet: Predicting remaining cataract surgery duration, 2021.

\bibitem{pucci2023}
Davide Pucci, Federico Becattini, and Alberto Del~Bimbo.
\newblock Joint-based action progress prediction.
\newblock {\em Sensors}, 23(1), 2023.

\bibitem{redmon2016}
Joseph Redmon and Ali Farhadi.
\newblock Yolo9000: Better, faster, stronger, 2016.

\bibitem{simonyan2015}
Karen Simonyan and Andrew Zisserman.
\newblock Very deep convolutional networks for large-scale image recognition,
  2015.

\bibitem{soomro2012}
Khurram Soomro, Amir~Roshan Zamir, and Mubarak Shah.
\newblock Ucf101: A dataset of 101 human actions classes from videos in the
  wild, 2012.

\bibitem{twinanda2016}
Andru~Putra Twinanda, Sherif Shehata, Didier Mutter, Jacques Marescaux, Michel
  de Mathelin, and Nicolas Padoy.
\newblock Endonet: A deep architecture for recognition tasks on laparoscopic
  videos, 2016.

\bibitem{twinanda2019}
Andru~Putra Twinanda, Gaurav Yengera, Didier Mutter, Jacques Marescaux, and
  Nicolas Padoy.
\newblock {RSDNet}: Learning to predict remaining surgery duration from
  laparoscopic videos without manual annotations.
\newblock {\em {IEEE} Transactions on Medical Imaging}, 38(4):1069--1078, apr
  2019.

\bibitem{vanamsterdam2020}
Beatrice van Amsterdam, Matthew~J. Clarkson, and Danail Stoyanov.
\newblock Multi-task recurrent neural network for surgical gesture recognition
  and progress prediction, 2020.

\bibitem{vidalmata2020}
Rosaura~G. VidalMata, Walter~J. Scheirer, Anna Kukleva, David Cox, and Hilde
  Kuehne.
\newblock Joint visual-temporal embedding for unsupervised learning of actions
  in untrimmed sequences, 2020.

\bibitem{wang2023}
Bowen Wang, Liangzhi Li, Yuta Nakashima, Ryo Kawasaki, and Hajime Nagahara.
\newblock Real-time estimation of the remaining surgery duration for cataract
  surgery using deep convolutional neural networks and long short-term memory,
  2023.

\bibitem{wang2013}
Heng Wang and Cordelia Schmid.
\newblock Action recognition with improved trajectories.
\newblock In {\em 2013 IEEE International Conference on Computer Vision}, pages
  3551--3558, 2013.

\bibitem{yengera2018}
Gaurav Yengera, Didier Mutter, Jacques Marescaux, and Nicolas Padoy.
\newblock Less is more: Surgical phase recognition with less annotations
  through self-supervised pre-training of cnn-lstm networks, 2018.

\end{thebibliography}
}

\appendix
% % \slp{move to appendix} 
% Figure ~\ref{fig:visualisation} shows an example with 3 training videos of length 10, 20, and 30, and a test video of length 50. We can see that the predicted progress follows an average trajectory, which changes at frame 10 and frame 20 when video 1 and video 2 end. The test video is 20 frames longer than the longest train video so after frame 30 the predicted progress stays pinned at $1.0$.
% % \slp{of what? say a bit more? what are the axis, what should we see?}. 
% Note, that as opposed to the progress of an activity this baseline is not guaranteed to be a monotonically non-decreasing progress.

% \begin{figure*}
% \begin{center}
%    \includegraphics[width=1.0\linewidth]{media/avg_index_example.pdf}
% \end{center}
%    \caption{Progress values of 3 videos of length 10, 20, and 30 (a) and their resulting average index baseline applied to a video of length 50 (b). The average index baseline follows the average trajectory of each video, changing directory at frame 10 and 20 when video 1 and video 2 end. After frame 30 the average index baseline predicts a constant progress value of $1.0$.
%    % \slp{Use only pdf figures, make the labels bigger and the line thicker. Explain the axes here in text. Conclude what should the reader see.}
%    }
% \label{fig:visualisation}
% \end{figure*}

\end{document}